%% file: main.tex
\def\year{2017}\relax
\documentclass[letterpaper]{article}
\usepackage{aaai17}
\usepackage{times}
\usepackage{helvet}
\usepackage{courier}

\usepackage{url}
\usepackage{verbatim}
\usepackage{color}
\usepackage{lipsum}
\usepackage{subfig}
\usepackage{graphicx}
\usepackage{tabularx}
\usepackage{paralist}

\newcommand{\TG}{\textit{target word}}

\makeatletter
\def\url@leostyle{%
  \@ifundefined{selectfont}{\def\UrlFont{\sf}}{\def\UrlFont{\small\bf\ttfamily}}}
\makeatother
\begin{document}
%
\title{Is this word borrowed? An automatic approach to quantify\\ the likeliness of borrowing in social media}

\author{$^1$Jasabanta Patro, $^2$Bidisha Samanta, $^3$Saurabh Singh, \\\textbf{$^4$Prithwish Mukherjee, $^5$Monojit Choudhury, $^6$Animesh Mukherjee}\\
$^{1,2,3,4,6}$ Indian Institute of Technology Kharagpur, India -- 721302 , $^5$Microsoft Research India, Bangalore -- 500033 \\ $^1$jasabantapatro, $^2$bidisha, $^3$saurabhsingh\}@iitkgp.ac.in, \\ $^4$pritspido@gmail.com, $^5$monojitc@microsoft.com, $^6$animeshm@cse.iitkgp.ernet.in \\
\\
}

\maketitle
\input{00Abstract}
\input{10Introduction}
\input{20Relatedwork}
\input{30Datasets}

\input{40Methodology}

\input{50Evaluation}

\input{60Conclusion}

\bibliographystyle{aaai}
\bibliography{main}
\end{document}

%% file: 00Abstract.tex
\begin{abstract}
\textit{Code-mixing} or {\em code-switching} refer to the phenomenon of effortless and natural switching between two or more languages in a single conversation, sometimes even in a single utterance, by multilingual speakers. However, use a foreign word in a language does not necessarily mean that the speaker is code-switching, because often languages {\em borrow} lexical items from other languages. 
If a word is borrowed, it becomes a part of the lexicon of a language; whereas, during code-switching the speaker is aware that the conversation involves multiple languages and often the switching is intentional. Identifying whether a non-native word used by a bilingual speaker is due to borrowing or code-switching is not only of fundamental importance to theories of multilingualism, but it is also an essential prerequisite towards development of language and speech technologies for multilingual communities. 

In this paper, we present for the first time, a series of computational methods to identify the likeliness of a word being borrowed or code-mixed, based on the signals from social media. In particular, we use tweets from English-Hindi bilinguals from India to predict word borrowing. We first propose a method to sample a set of candidate words from the social media data using a context based clustering approach. Next, we propose three novel and similar metrics based on the usage of these words by the users in different tweets; we then apply these metrics to score and rank the candidate words indicating their likeliness of being borrowed. We compare these rankings with a ground truth ranking constructed through a human judgement experiment. The Spearman's rank correlation between the two rankings ($\sim 0.62$ for all the three metric variants) is more than double the value (0.26) of the most competitive existing baseline reported in the literature. Some other striking observations are -- (i) the correlation is higher for the ground truth data elicited from the younger participants (age $<$ 30) than that from the older participants; since language change is brought about by the younger generation, this possibly indicates that social media is able to provide very early signals of borrowing, and (ii) those participants who use mixed-language for tweeting the least, provide the best signals of borrowing. 
 

\end{abstract}

    

%% file: 10Introduction.tex
\section{Introduction}
In multilingual societies, where two or more languages co-exist and are used regularly by the users, several language-interaction phenomena are observed. {\em Code-switching} or {\em code-mixing} is one such common phenomenon where speakers spontaneously switch between multiple languages in a single conversation, sometime even within a single sentence or phrase~\cite{auer}. A related but linguistically and cognitively distinct phenomenon is {\em lexical borrowing} (or simply, {\em borrowing}), where a word or phrase from a foreign language is used as a part of the native vocabulary of a language. Examples of word borrowing are widespread; for instance, in Dutch the English word ``sale'' is now used more frequently than the Dutch equivalent ``uitverkoop''. Some English words like ``shop'' are even inflected in Dutch as ``shoppen'' and heavily used. Similar examples of borrowing of words from English can be found across many other languages like Spanish, German, French, Chinese etc. The opposite is also true, i.e., words like ``tortilla'', ``tequila'', ``ramedan'', ``couscous'' and ``tandoori'' are borrowed in English from other languages. 

While it is difficult in general to ascertain whether a foreign word or phrase used in an utterance is borrowed or just an instance of code-mixing~\cite{sharma2014borrowing}, one tell tale sign is that only proficient multilinguals can code-mix, while even monolingual speakers can use borrowed words because, by definition, these are part of the vocabulary of a language. In other words, just because an English speaker understands and uses the word ``tortilla'' does not imply that she can speak or understand Spanish. In this work, we develop a novel method to identify borrowed words, or more formally the likeliness that a word is in the process of getting borrowed. 

A borrowed word from a foreign language, initially appears frequently in speech, then gradually in print media like newspaper and finally it loses its origin's identity and is used in the native language resulting in an inclusion in the dictionary of the native language~\cite{myers2002contact,thomason2003contact}. 
However, early-stage automatic identification of whether a word is likely to be borrowed is known to be a hard problem. The main hurdles are (i) the lack of well-defined linguistic signals of borrowing, primarily because it is a socio-linguistic phenomenon closely related to acceptability and frequency, (ii) borrowing is a dynamic process; new borrowed words enter the lexicon of a language as old words, both native and borrowed, might slowly fade away from usage, and (iii) it is a population level phenomenon that necessitates data from a large portion of the population unlike standard natural language corpora that typically comes from a very small set of authors.


The above reasons motivate us to resort to the social media (in particular, Twitter), where a large population of bilingual/multilingual speakers are known to often tweet in code-mixed colloquial languages~\cite{carter2013microblog,solorio,vyas2014pos}. Fig.~\ref{fig:ex} shows some typical tweets from Hindi-English bilinguals from India. 

The central hypothesis of this study is as follows: Since, language use over social media is informal, has speech-like characteristics~\cite{carmen12} and involves a large population of speakers from a wide range of socio-linguistic communities, it should be possible to extract early signals of likeliness of borrowing of a word from the language usage patterns from social media data. In this study, we analyze the tweets of the English-Hindi bilinguals from India for predicting the likeliness of borrowing. Note that our approach and the metrics that we propose can be easily generalised for any other language pair. Hindi is assumed to be the native language while English is the foreign language from where a candidate word could be borrowed. We present some typical examples of such bilingual tweets in figure~\ref{fig:ex}.
\vspace{2mm}

\begin{figure}
\caption{Some example code-mixed tweets from English-Hindi bilinguals. Hindi words are in italics.}
\fbox{\begin{centering}\begin{minipage}{0.43\textwidth}
{\tt Huge traffic restrictions for PM's visit to \#blast site mean deserted roads in \#Hyderabad. \textit{``Itna sanaata kyon hai bhai?''}}\\
{\bf Translation}: Huge traffic restrictions for Prime Minister's visit to the blast site mean deserted roads in Hyderabad. ``Why is there so much silence, bro?''\\
\\
{\tt MMS  will go to \#HyderabadBlast site to take \textit{jayeja} of area \& say \textit{Hazaaron Jawabon Se Acchi Hai Meri Khamoshi} \#ThikHai}\\
{\bf Translation}: MMS (name of a politician) will go to \#HyderabadBlast site to take a survey of the area and say "My silence is better than a thousand answers." \#ThikHai
\end{minipage}
\end{centering}
}\label{fig:ex}
\end{figure}
\vspace{2mm}

The main stages of our research are as follows:

\begin{compactitem}
\item \textit{Unsupervised method for selecting candidate words from a large corpora}: We propose a context based clustering approach to appropriately sample a list of candidate English words from the bilingual tweets to investigate the borrowing phenomenon.   
\item \textit{Ground truth generation}: We launch an extensive survey among $58$ human judges of various age groups and various educational backgrounds to collect responses indicating if each of the candidate English word is likely borrowed. Further, we aggregate the responses to rank the candidate words based on the likeliness of borrowing.
\item \textit{Metric to quantify the likeliness of borrowing from social media signals}: We define three novel and closely similar metrics that serve as social signals indicating the likeliness of borrowing. All the three metrics attempt, in some form, to estimate the extent to which users use a candidate English word in an otherwise (predominantly) Hindi tweet. The higher this extent is for a candidate word, the higher should be the likeliness of borrowing. 
\item \textit{Experiments}: We compare the likeliness of borrowing as predicted by our model and a baseline model with that of ground-truth obtained from human judges.
\end{compactitem}
Finally, our key results are outlined below:
\begin{compactitem}
\item The Spearman's rank correlation between the ground-truth ranking and the ranking based on our metric is $\sim 0.62$ for all the three variants. Remarkably, this value is more than double the value (0.26) if we use the most competitive baseline~\cite{sharma2014borrowing} reported in the literature for ranking the words.
\item In case the candidate words are surely instances of borrowing (i.e., are at the top of the ground truth ranking), our metrics does as good as the baseline. However as one moves down the rank list, our metric overwhelmingly outperforms the baseline.
\item Interestingly, the responses of the judges in the age group below $30$ seem to correspond even better with our metrics. Since language change is brought about mostly by the younger population, this might possibly mean that our metrics are able to capture the early signals of borrowing.
\item Those users that mix languages the least in their tweets present the best signals of borrowing in case they do mix the languages (correlation of our metrics estimated from the tweets of these users with that of the ground truth is $\sim 0.65$). 
\end{compactitem}

Note that apart from the fact that this is a hard but very interesting socio-linguitic problem (which was our primary motivation to choose this problem), the ability to automatically distinguish cases of code-borrowing from code-mixing can have a strong impact on engineering applications like multilingual information retrieval and natural language processing tasks. For instance, if a query has more than one language, say $L_1$ \& $L_2$, and if all the words of $L_2$ are borrowed in $L_1$ and not vice versa, then we can infer that it is possibly an $L_1$ query, thus giving priority to only $L_1$ documents. However, if $L_1$ and $L_2$ do not have mutually borrowed words, then it is a truly multilingual query and so, we should retrieve $L_1$, $L_2$ and $L_1$-$L_2$ mixed documents for that query. Furthermore, several studies have brought up into notice the prevalence of multilinguality, and more specifically use of code-mixing in user-generated content on social media~\cite{sharma2014borrowing,barman2014code,solorio}. Processing of such content, for example sentiment and opinion detection of code-mixed tweets~\cite{rudra16}, requires language detection, and more specifically language-switch detection because language switch often signals a change in opinion or sentiment. However, as argued in the case of IR, borrowed words do not necessarily indicate language switch, even though they belong to a different language.

%% file: 20Relatedwork.tex

\section{Related Work}

\noindent{\bf Early literature}: In linguistics, \textit{code-mixing} and \textit{word-borrowing} are often studied under the broader scope of language evolution and change. Both of these phenomena are observed when the speakers of two or more languages come in close contact with one another. Linguists have for a long time focused on the sociological and the conversational necessity of borrowing and mixing~\cite{auer,muysken}. 

\noindent{\bf Automatic processing of mixed text}: One of the first works in automatic processing of code-mixed text dates more than thirty years back~\cite{joshi}, while the task of automatic language identification is even earlier~\cite{gold}. Since then there have been many works in this area. For instance,~\cite{sankoff1990case} reported the complexity of choosing features to determine the indicators of borrowing. This work further shows that it is not always true that only highly frequent words are borrowed, nonce words could also be borrowed along with the frequent words. According to~\cite{field2002linguistic}, the principle of system compatibility/incompatibility can be used to determine the nature of systematic interactions of human beings.~\cite{nzai2014understanding} analyzed the formal conversation of Spanish-English multilingual people and found that code mixing/ word borrowing is not only restricted to daily speech but also in formal conversations.~\cite{hadei2016single} showed that phonological integration could be evaluated to understand the phenomenon of word borrowing. In similar lines,~\cite{sebonde2014code} showed morphological and syntactic features could be good indicators for numerical borrowings.~\cite{senaratne2013borrowings} reported that in many languages English words are likely to be borrowed in both formal and semi-formal text.

In parallel, works on language identification have also gained a lot of attention; for instance,~\cite{cardenas2009code,monojit2011challenges,prager1999linguini} have analyzed the problems of language identification in code-mixed text. 

\noindent{\bf Mixing in short texts and emails}: There have also been works studying code-mixing in short texts like SMS.~\cite{sotillo} investigated various types of code-mixing in a corpora of 880 SMS text messages. The author observed that most often mixing takes place at the beginning of a sentence as well as through simple insertions. Similar observations about chat messages have been reported in~\cite{bock}.~\cite{negron} reported a study on code-mixing in the emails of five Spanish-English bilinguals and reiterated similar results as in the case of SMS and chat messages. However, studies of code-mixing with Chinese-English bilinguals from Hong Kong~\cite{li} and Macao~\cite{hksan} brings forth results that contrasts the aforementioned findings and indicate that in these societies code-mixing is driven more by linguistic than social motivations. 

\noindent{\bf Mixing in social media}: Recently, the advent of social media has immensely propelled the research on code-mixing and word-borrowing as dynamical social phenomena. Such research has been inspired by many interesting studies on various general phenomena on language dynamics~\cite{mizil,kulkarni}. 

~\cite{hidayat} noted that in Facebook, users mostly preferred inter-sentential mixing and showed that 45\% of the mixing originated from real lexical needs, 40\% was used for conversations on a particular topic and the rest 5\% for content clarification. In contrast,~\cite{das} showed that in case of Facebook messages, intra-sentential mixing accounted for more than half of the cases while inter-sentential mixing accounted only for about one-third of the cases. There have also been quite a few studies on Twitter; for instance~\cite{carter} collected tweets from five different languages (Dutch, English, French, German, and Spanish), and manually inspected the multilingual micro-blogs to identify the dominant language for a specific tweet. Subsequently, a character $n$-gram distance metric was introduced to automate the process. In fact, in the First Workshop on \textit{Computational Approaches to Code Switching} a shared task on code-mixing in tweets was launched and four different code-mixed corpora were collected from Twitter as a part of the shared task~\cite{solorio}. Language identification task has also been handled for English-Hindi and English-Bengali code-mixed tweets in~\cite{das13}. Part-of-speech tagging have been recently done for code-mixed English-Hindi tweets~\cite{vyas2014pos}.

Despite the presence of such a huge literature, there has not been much attempt to quantify the likeliness of borrowing of candidate foreign word in a native language. The only work that makes an attempt in this direction is~\cite{sharma2014borrowing}.
The biggest hurdle in doing such a study is that borrowing is a strongly social phenomenon and it is difficult to identify suitable indicators of such a lexical diffusion process unless one has access to large population level data. In this work, we show for the first time how certain simple and closely related signals encoding the language usage of social meida users can help us construe appropriate metrics to quantify the likeliness of borrowing of a foreign word. We show, through a series of rigorous experiments how our metrics by far beats the most competitive baseline reported in the literature.

%% file: 30Datasets.tex
\section{Datasets and preprocessing}

We consider English-Hindi code-mixed tweets from India for the purpose of our experiments. To bootstrap the data collection process, we crawl tweets (between Nov 2015 and Jan 2016) related to 28 hashtags representing different Indian contexts covering important topics such as sports, religion, movies, politics etc. This process results in the collection of 811981 tweets. We language-tag (see details later in this section) each tweet so crawled and find that there are 3982 users who use mixed language for tweeting. We then systematically crawl the time lines of these 3982 users between Feb 2016 and March 2016 to gather more mixed language tweets. Using this two step process we collect a total of 1550714 distinct tweets. From this data, we filter out tweets that are not written in romanized script, tweets having only URLs and tweets having empty content. Post filtering we obtain 787606 tweets which we use for the rest of the analysis.

\noindent{\bf Language tagging}: We tag each word in a tweet with the language of its origin using the method outlined in~\cite{gella2013query}. The different tags that a word can have are: \textit{En} (English), \textit{Hi} (Hindi), \textit{NE} (Named Entity) and \textit{Others} (hash-tags, URLs, twitter user names etc.). Based on the word level tag, we create a tweet level tag as follows: 

\begin{compactenum}
	\item \textit{En}: Almost every word ($> 90\%$) in the tweet is tagged as \textit{En}.
    \item \textit{Hi}: Almost every word ($> 90\%$) in the tweet is tagged as \textit{Hi}.
    \item \textit{CME}: Code-mixed tweet but majority (i.e., $>50\%$) of the words are tagged as \textit{En}.
	\item \textit{CMH}: Code-mixed tweet but majority (i.e., $>50\%$) of the words are tagged as \textit{Hi}.
	\item \textit{CMEQ}: Code-mixed tweet having equal number of words tagged as \textit{En} and \textit{Hi} respectively. 
	\item \textit{Code Switched}: There is a trail of Hindi words followed by a trail of English words or vice versa.
\end{compactenum}

In table~\ref{tab:stat} we note the number and percentage of tweets in each of the above six categories in which the tweets are labeled. Like the word level, the tagger also provides a phrase level language tag. Once again, the different tags that an entire phrase can have are: \textit{En}, \textit{Hi} and \textit{Oth} (Other). We shall use these word and phrase level tags in order to define our metrics in the next section.

\noindent{\bf Frequent foreign words}: In this step we compute the most frequent foreign (i.e., English words) in our tweet corpus. Since we are interested in the frequency of the English word only when it appears as a foreign word we do not consider the (i) \textit{Hi} tweets since they do not have any foreign word, (ii) \textit{En} tweets since here the English words are not foreign words and the (iii) code-switched tweets. Based on the frequency of usage of English as a foreign word, we select the top 1000 English words. Removal of stop words and text normalization leaves beyond 230 nouns. We note these words in the box below:

\fbox{\begin{minipage}{0.43\textwidth}\scriptsize
\textit{`welfare', `anniversary', `tribute', `box', `victory', `thing', `lot', `youth', `need', `nation', `birth', `people', `muslims', `god', `water', `teacher', '`airport', `army', `room', `answer', `blood', `law', `light', `chief', `green', `office', `border', `food', `university', `side', `event', `health', `reason', `city', `station', `theatre', `crore', `ground', `college', `bomb', `corruption', `court', `opposition', `respect', `life', `air', `rail', `student', `government', `mom', `aunty', `weekend', `age', `protest', `guy', `company', `bollywood', `place', `message', `friend', `mind', `mobile', `view', `volunteer', `moment', `rest', `suicide', `lyrics', `group', `death', `home', `way', `brother', `house', `blue', `wedding', `reaction', `terrorist', `person', `mother', `press', `election', `power', `question', `lord', `birthday', `president', `half', `day', `internet', `number', `service', `morning', `waste', `voice', `evening', `night', `luck', `son', `favourite', `captain', `video', `sun', `body', `experience', `family', `use', `music', `date', `teaser', `share', `man', `paper', `lunch', `logo', `season', `job', `game', `post', `gift', `poster', `film', `test', `performance', `price', `plan', `class', `shot', `report', `prime', `state', `exam', `success', `road', `form', `problem', `check', `wife', `boy', `car', `heart', `scam', `style', `police', 'issue', 'card', 'country', 'boss', 'party', 'entry', 'uncle', 'politics', `father', `parliament', `work', `sunday', `story', `play', `request', `week', `playlist', `matter', `superstar', `traffic', `suit', `woman', `cool', `history', `money', `bat', `seat', `score', `photo', `parents', `decision', `girlfriend', `picture', `month', `song', `word', `school', `hero', `degree', `love', `train', `end', `wrong', `main', `scene', `bank', `miss', `king', `channel', `face', `link', `news', `media', `mood', `book', `selfie', `bus', `status', `petrol', `railway', `budget', `well', `development', `team', `phone', `baby', `sir', `interview', `fan', `trailer', `year', `girl', `time', `review', `madam', `movie', `minister', `joke', `century', `cup', `match', `world', `temple', `wicket', `cricket', `star'}
\end{minipage}}

\noindent{\bf Newspaper dataset}: As we shall see, for the construction of the baseline ranking we shall need to resort to counting the frequency of the foreign words (i.e., English words) and their Hindi translations in a newspaper corpus as has been outlined in~\cite{sharma2014borrowing}. For this purpose, we use the FIRE dataset built from the Hindi Jagaran newspaper corpus\footnote{Jagaran corpus: \url{http:/fire.irsi.res.in/fire/static/data}} which is written in Devanagari script. 

\begin{table}
\centering
\caption{Number and percentage of tweets in each of the six categories in which the tweets are labeled.}
\label{tab:stat}
\scalebox{0.8}{
\begin{tabular}{|l|l|l|l|}
\hline 
Type & Number of Tweets & percentage \\ \hline \hline
\textit{En}  & 645655 & 81.97 \\ \hline
\textit{Hi}   & 24960 &  3.16 \\ \hline
\textit{CME}  & 31998  & 4.06 \\ \hline
\textit{CMH}  & 39877 & 5.06 \\ \hline
\textit{CMEQ} & 3584    & 0.455 \\ \hline
\textit{CS} & 41532  & 5.27 \\ \hline
\end{tabular}
}
\end{table}

%% file: 40Methodology.tex
\section{Methodology}
In this section, we present the baseline metric and our metrics based on the population level signals to quantify the likeliness of borrowing. We also present a scheme for sampling an appropriate set of target words  for the purpose of our experiments.

\subsection{Baseline metric and ranking}
\noindent{\em Baseline metric} -- We consider the $log(\frac{F_E}{F_H})$ value proposed in~\cite{sharma2014borrowing} as the baseline metric. Here $F_E$ denotes the frequency of the Devanagari transliterated form of the word $w$ in the Jagaran corpus. $F_H$, on the other hand, denotes the frequency of the Hindi translation of the word $w$ in the Jagaran corpus. Both the transliteration and the translation of the words have been done by a set of volunteers who are native Hindi speakers. The authors in~\cite{sharma2014borrowing} claim that the more positive the value of this metric is for a word $w$, the higher is the likeliness of its being borrowed. The more negative the value is, the higher are the chances that the word $w$ is an instance of code-mixing. 

\noindent {\em Ranking} -- Based on the values obtained from the above metric for a set of target words, we rank these words; words with high positive values feature at the top of the rank list and words with high negative values feature at the bottom of the list. For two words having the same $log(\frac{F_E}{F_H})$ value, we resolve the conflict by assigning each of these the average of their two rank positions. In the next section, we shall compare this rank list with the one obtained from the ground truth responses.

\subsection{Proposed metric and ranking}
In this section, we present three novel and closely related metrics based on the language usage patterns of the users of social media.  

\noindent{\em Unique User Ratio} ($UUR$) -- The Unique User Ratio for word usage across languages is defined as follows: 
\begin{equation}
UUR(w) = \frac{U_{Hi}+U_{CMH}}{U_{En}}
\end{equation}
where $U_{Hi}$ is the number of unique users who have used the word $w$ in a Hindi tweet at least once, $U_{En}$ is the number of unique users who have used the word $w$ in an English tweet at least once and $U_{CMH}$ is the number of users who have used the word $w$ in a code-mixed Hindi tweet at least once. Higher the value of $UUR$ higher should be the likeliness of the word $w$ being borrowed.
 
\noindent{\em Unique Tweet Ratio} ($UTR$) -- The Unique Tweet Ratio for word usage across languages is defined as follows: 
\begin{equation}
UTR(w) = \frac{T_{Hi}+T_{CMH}}{T_{En}}
\end{equation}
where $T_{Hi}$ is the total number of Hindi tweets which contain the word $w$, $T_{En}$ is the total number of English tweets which contain the word $w$ and $T_{CMH}$ is the total number of CMH tweets which contain the word $w$ . Higher the value of $UTR$ higher should be the likeliness of the word $w$ being borrowed.

\noindent{\em Unique Phrase Ratio} ($UPR$) -- The Unique Phrase Ratio for word usage across languages is defined as follows: 
\begin{equation}
UPR(w) = \frac{P_{Hi}}{P_{En}}
\end{equation}
where $P_{Hi}$ is the number of Hindi phrases which contain the word $w$, $P_{En}$ is the number of English phrases which contain the word $w$. Note that unlike the definitions of $UUR$ and $UTR$ that exploit the word level language tags, the definition of $UPR$ exploits the phrase level language tags. Once again, higher the value of $UPR$ higher should be the likeliness of the word $w$ being borrowed.

\noindent{\em Ranking} -- We prepare a separate rank list of the target words based on each of the three proposed metrics -- $UUR$, $UTR$ and $UPR$. In the next section, we shall compare these rank lists with the one prepared from the ground truth responses.  

\subsection{Target word selection}
We use following method to select the final set of \TG\textit{s} from the 230 nouns for our evaluation. 
\subsubsection{Clustering}
In language processing, context plays an important role in understanding different properties of a word. For our study, we also attempt to use the language tags as features of the context words for a given \TG. Our hypothesis here is that there should exist classes of words that have similar context features and the likelihood of being borrowed in each class should be different. For example, when an English word is surrounded by mostly Hindi words it seems to be more likely borrowed. We present two examples in the box below to illustrate this. 
\vspace{2mm} 
\fbox{\begin{centering}\begin{minipage}{0.43\textwidth}
{\em Example I}:\\
@KapilianPooja Welcome. \textbf{\textit{Film}} \textit{jaroor dekhna. Nahi to injection ready hai.} \\ 
- Dr Mashoor Gulati (@ItsDrGulati)\\
{\em Example II}:\\
@HelpU\_Trust @DrKumarVishwas\\
\textit{Kuch to ache se karo sirji}....\\
\textit{Har jagah bhaagte rehna} is not a good \textbf{\textit{thing}}.\\
- lovely sethi(@lovelysethii)
\end{minipage}
\end{centering}
}
\vspace{2mm}

In \textit{Example I} the English word ``film'' is surrounded by mostly Hindi words. On the other hand, in \textit{Example II} the English word ``thing'' is surrounded mostly by English words. Note that the word ``film'' is very commonly used by Hindi monolingual speakers and is therefore highly likely to have been borrowed unlike the English word ``thing'' which is arguably an instance of mixing. This socio-linguistic difference seems to be very appropriately captured by the language tag of the surrounding words of these two words in the respective tweets.

Therefore, as a first step toward sampling a set of representative \TG\textit{s}, we cluster the list of 230 words into contextually similar groups as follows.

\noindent{\em Construction of feature vectors} -- We represent a context feature for a \TG~as a tuple $\{P_b, P_a\}$ where $P_b$ is the language tag for the word before the \TG~(i.e., the left context) and $P_a$ is the language tag of the word after the \TG~(i.e., the right context). Each of $P_b$ and $P_a$ can be either ``E'' indicating English, ``H'' indicating Hindi or ``\$'' indicating the boundary (i.e., beginning or end) of the tweet. Thus, we have \textit{eight} feature combinations of the left and the right contexts of a \TG~-- ``EE'', ``HH'', ``EH'', ``HE'', ``\$E'', ``E\$'', ``\$H'', ``H\$'' while ``\$\$'' is not possible. For every \TG, we compute the percentage of occurrences of each of these combinations.   

\begin{figure}
\caption{Stacked plot representing feature vectors of four different words. Note that the feature vectors of the word pairs (i) ``job'' and ``film'' and (ii) ``moment'' and ``protest'' are very similar. The fractional counts of the eight combinations for each tweet category should sum up to one; since there are three tweet categories so the total size of the stacked plot is three.}
\centering
\vspace{-3mm}
\includegraphics[width=0.49\textwidth]{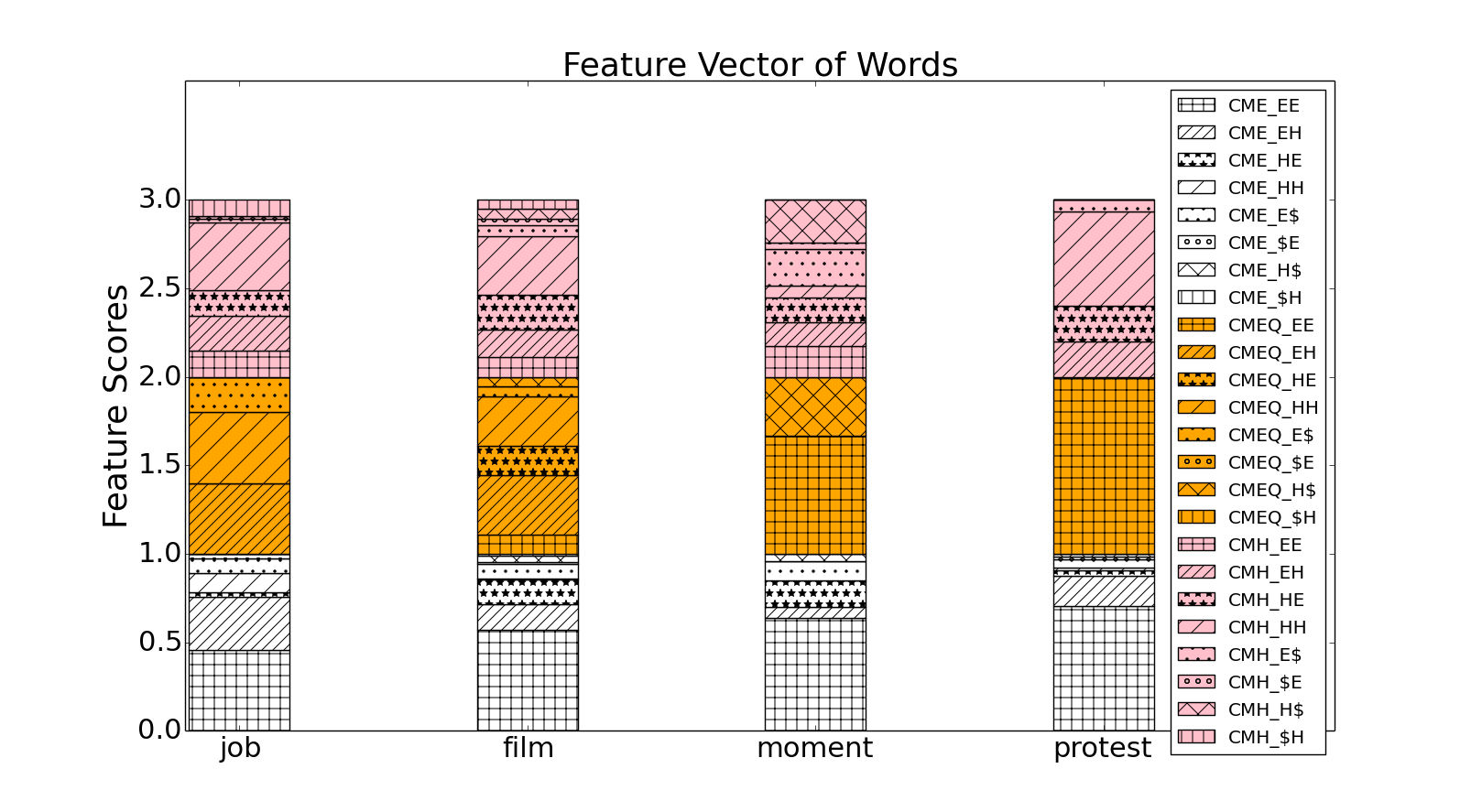}
\label{fig:24Feature}
\vspace{-4mm}
\end{figure}

Note that we compute these percentages from the three different categories of tweets -- \textit{CME}, \textit{CMH} and \textit{CMEQ}. Thus, for every \TG~we have a final feature vector of length 24, each entry denoting the percentage of one feature combination in a particular tweet category. We show example feature vectors for some words in figure~\ref{fig:24Feature}.

\noindent{\em $K$-means clustering} -- We use the feature representation of the words to cluster them into contextually similar groups. We use $K$-means clustering~\cite{hartigan1979algorithm} for the purpose of clustering. Hence we vary the value of $K$ and using the traditional elbow method~\cite{tibshirani2001estimating} we obtain 15 as the optimal value of $K$. This process therefore groups the 230 nouns into 15 different clusters. 

\subsubsection{Final selection of target words}
In this final stage, we select two sets of target words using two different strategies as follows.

\noindent{\em Baseline biased words} ($bbw$) -- We compute $log(\frac{F_E}{F_H})$ values for all the words in each cluster. From each cluster, we select two words having the highest and the lowest values of $log(\frac{F_E}{F_H})$. This constitutes a set of 30 words. Note that this selection is biased toward the baseline metric discussed above so as to give maximum possible advantage to the baseline ranking. 

\noindent{\em Randomly selected words} ($ran$) -- In order to check how well all the proposed metrics behave in general, i.e., to investigate the recall of each metric, we randomly select 27 words from the 15 clusters. These 27 words are kept completely different from the baseline biased words.

\noindent{\em Full set of words} ($full$) -- Thus, in total we select 57 target words for the purpose of our evaluation. We present these words in the box below.
  
\fbox{\begin{minipage}{0.43\textwidth}\small
\textit{Baseline biased words} -- \textit{'thing', 'way', 'woman', 'press', 'wrong', 'well', 'matter', 'reason', 'question', 'guy', 'moment', 'week', 'luck', 'president', 'body', 'job', 'car', 'god', 'gift', 'status', 'university', 'lyrics', 'road', 'politics', 'parliament', 'review', 'scene', 'seat', 'film', 'degree'}\\
\textit{Randomly selected words} -- \textit{ 'people', 'play', 'house', 'service', 'rest', 'boy', 'month', 'money', 'cool', 'development', 'group', 'friend', 'day', 'performance', 'school', 'blue', 'room', 'interview', 'share', 'request', 'traffic', 'college', 'star', 'class', 'superstar', 'petrol', 'uncle'}
\end{minipage}}


%% file: 50Evaluation.tex
\section{Results and Discussion}\label{sec:eval}

\subsection{Evaluation criteria}
In this section, we present extensive evaluation to demonstrate the effectiveness of the three proposed metrics $UUR$, $UTR$ and $UPR$ in quantifying the likeliness of borrowing. In particular, we present a four step approach as follows. We measure (i) how well the $UUR$, $UTR$ and $UPR$ based ranking of the $bbw$ set, the $ran$ set and the $full$ set correlate with the ground truth ranking (discussed in the next section) in comparison to the rank given by the baseline metric, (ii) how well the different rank ranges obtained from our metric align with the ground truth as compared to the baseline metric, (iii) whether there are some systematic effects of the age group of the survey participants on the rank correspondence, and (iv) how our metrics if computed from the tweets of users who (a) rarely mix languages, (b) almost always mix languages and (c) are in between (a) and (b), align with the ground truth. 
 
\noindent{\bf Rank correlation}: We measure the standard Spearman's rank correlation ($\rho$)~\cite{zar1972significance} pairwise between rank lists generated by (i) $UUR$ and ground truth, (ii) $UTR$ and ground truth, (iii) $UPR$ and ground truth, and (iv) baseline and ground truth.  

We shall describe the next three measurements taking $UUR$ as the running example. The same can be extended verbatim for the other two similar metrics.

\noindent{\bf Rank ranges}: We split each of the three rank lists ($UUR$, ground truth and baseline) into five different equal-sized ranges as follows -- (i) surely borrowed (SB) containing top 20\% words from each list, (ii) likely borrowed (LB) containing the next 20\% words from each list, (iii) borderline (BL) constituting the subsequent 20\% words from each list, (iv) likely mixed (LM) comprising the next 20\% words from each list and (v) surely mixed (SM) having the last 20\% words from each rank list. Therefore, we have three sets of five buckets, one set each for $UUR$, the ground truth and the baseline based rank list. 

Next we calculate the bucket-wise correspondence between (i) the $UUR$ and the ground truth set and (ii) the baseline and the ground truth set in terms of standard \textit{precision} and \textit{recall} measures. For our purpose, we adapt these measures as follows.\\
$G$: ground truth bucket set, $B_b$: baseline bucket set, $U_b$: $UUR$ bucket set;\\
$BS \in \{B_b, U_b\}$, $T$ (type of bucket) = \{SB, LB, BL, LM, SM\};\\
$b_t$ = words in type $t$ bucket from $BS$, $g_{t}$ = words in type $t$ bucket from $G$, $t \in T$;\\ 
$tp_{t}$ (no. of true positives) = $|b_{t} \cap g_{t}|$, $fp_{t}$ (no. of false positives) = $|b_{t} - g_{t}|$, $tn_{t}$ (no. of true negatives) = $|g_{t} - b_{t}|$; \\ 
Bucket-wise \textit{precision} and \textit{recall} are then given by:\\
$precision(b_{t}) = \frac{tp_{t}}{fp_{t} + tp_{t}}$ \\
$recall(b_{t}) = \frac{tp_{t}}{tn_{t} + tp_{t}}$\\
For a given set, we obtain the overall \textit{macro precision} (\textit{recall}) by averaging the \textit{precision} (\textit{recall}) values over the five buckets. For a given set, we also obtain the overall \textit{micro precision} by first adding the true positives across all the buckets and then normalizing by the sum of the true and the false positives over all the buckets. We take an equivalent approach for obtaining the \textit{micro recall}.

\noindent{\bf Age group effect}: Here we construct two ground truth rank lists one using the responses of the participants with age below 30 (young population) and the other using the responses of the rest of the participants (elderly population). Next we repeat the above two evaluations considering each of the new ground truth rank lists. 

\noindent{\bf Extent of language mixing}: Here we divide all the 3982 users into three categories -- (i) High (users who have more than 20\% of tweets as code-mixed), (ii) Mid (users who have 7--20\% of their tweets as code-mixed, and (iii) Low (users who have less than 7\% of their tweets as code-mixed). We create three $UUR$ based rank lists for each of these three user categories and respectively compare them with the ground truth rank list.

\subsection{Ground truth preparation}

Since it is very difficult to obtain a suitable ground truth to validate the effectiveness of our proposed ranking scheme, we launched an online survey to collect human judgment for each of the 57 \TG\textit{s}. 

\subsubsection{Online survey}
We conducted the online survey\footnote{Survey portal: \url{https://goo.gl/forms/L0kJm8BNMhRj0jA53}} among 58 volunteers majority of whom were either native Hindi speakers or had very high proficiency in reading and writing Hindi. The participants were selected from different age groups and different educational backgrounds. Every participant was asked to respond to a multiple choice question about each of the 57 \TG\textit{s}. Therefore, for every single \TG, 58 responses were gathered. The multiple choice question had the following three options and the participants were asked to select the one they preferred the most and found more natural -- (i) a Hindi sentence with the \TG~as the only English word, (ii) the same Hindi sentence in (i) but with the \TG~replaced by its Hindi translation and (iii) none of the above two options. There were no time restrictions imposed while gathering the responses, i.e., the volunteers theoretically had unlimited time to decide their responses for each \TG. 

\subsubsection{Language preference factor}
For each \TG, we compute a \textit{language preference factor} ($LPF$) defined as $(Count_{En}-Count_{Hi})$, where $Count_{Hi}$ refers to the number of survey participants who preferred the sentence containing the Hindi translation of the \TG~while $Count_{En}$ refers to the number of survey participants who preferred the sentence containing the \TG~itself. More positive values of $LPF$ denotes higher usage of \TG~as compared to its Hindi translation and therefore higher likeliness of the word being borrowed.

\subsubsection{Ground truth rank list generation}
We generate the ground truth rank list based on the $LPF$ score of a \TG. The word with the highest value of $LPF$ appears at the top of the ground truth rank list and so on in that order. Tie breaking between \TG\textit{s} having equal $LPF$ values is done by assigning average rank to each of these words.

\noindent{\bf Age group based rank list}: As discussed in the previous section, we prepare the age group based rank lists by first splitting the responses of the survey participants in two groups based on their age -- (i) young population (age $< 30$) and (ii) elderly population (age $\ge 30$). For each group we then construct a separate $LPF$ based ranking of the \TG\textit{s}.

\begin{figure*}[ht]
\caption{The rank ordered histograms of \TG\textit{s} ranked by various metrics.}
\centering
\vspace{-3mm}
\includegraphics[width=0.75\textwidth]{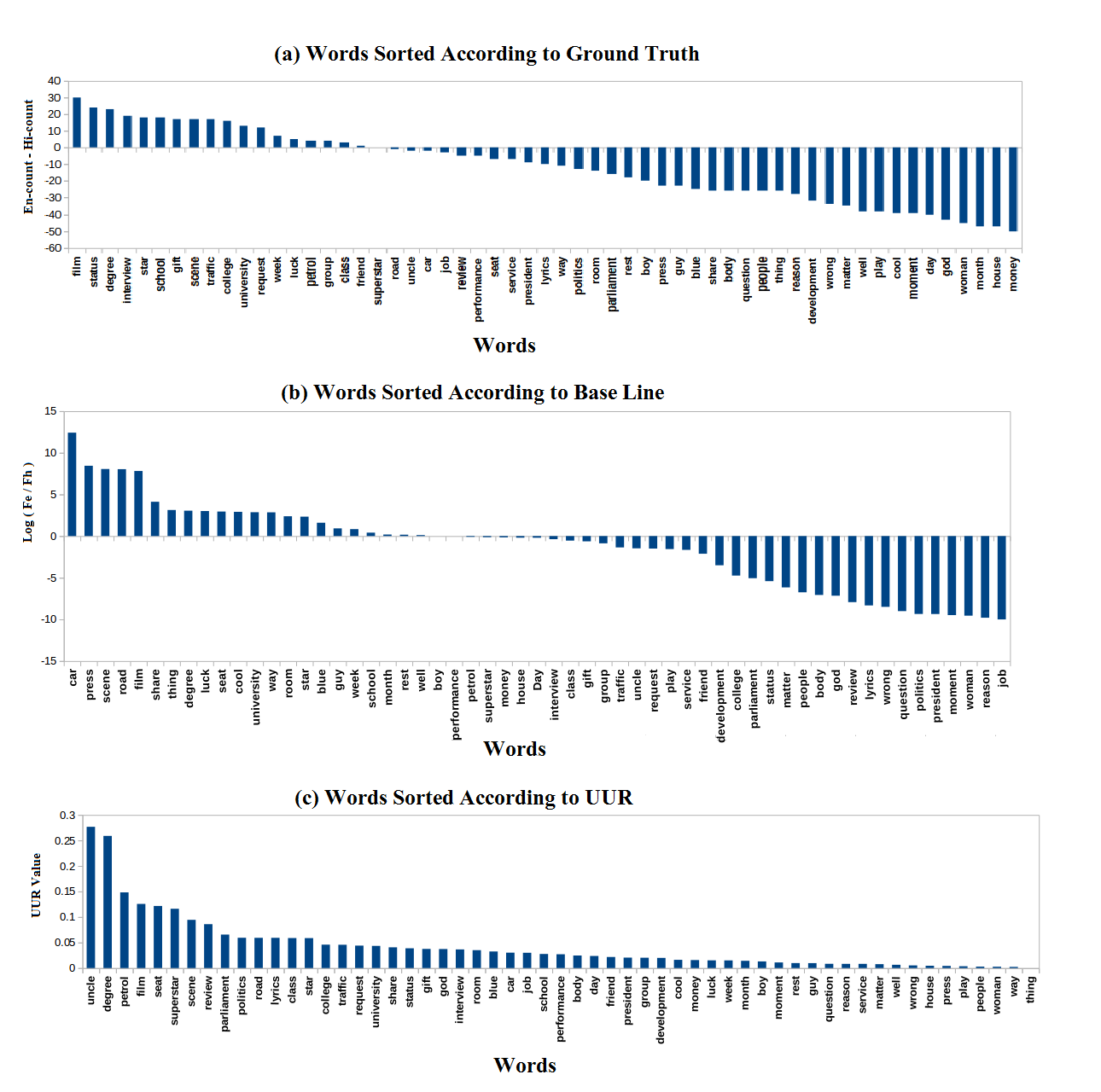}
\label{fig:All_List}
\end{figure*}

\subsection{Correlation among rank lists}
The Spearman's rank correlation coefficient ($\rho$) of the rank lists for the $bbw$ set, the $ran$ set and the $full$ set according to the baseline metric, $UUR$, $UTR$ and $UPR$ with respect to the ground truth metric $LPF$ are noted in~\ref{tab:SpearmanValue}. We observe that for the $full$ set, the $\rho$ between the rank lists obtained from all the three metrics $UUR$, $UTR$, and $UPR$ with respect to the ground truth is $\sim 0.62$ which is more than double the $\rho$ ($\sim 0.26$) between the baseline and the ground truth rank list. This clearly shows that the proposed metrics are able to identify the likeliness of borrowing quite accurately and far better than the baseline. Further, a remarkable observation is that our metrics outperform the baseline metric even for the $bbw$ set that is baseline-biased. Likewise, for the $ran$ set, our metrics outperform the baseline indicating a superior recall on arbitrary words. 

In order to help the readers to better visualize, we show in figure~\ref{fig:All_List} the ranking of the $full$ set of words obtained from the ground truth, the baseline and the $UUR$ metric. The figure clearly shows that $UUR$ correlates with the ground truth far better than the baseline. Very similar results are obtained for the two other mertics $UTR$ and $UPR$ and therefore not shown.

We present the subsequent results for the $full$ set and the $UUR$ metric. The results obtained for the other two metrics $UTR$ and $UPR$ are very similar and therefore not shown.

\begin{table}
\centering
\caption{Spearman's rank correlation coefficient ($\rho$) among the different rank lists. Best result is marked in bold.}
\label{tab:SpearmanValue}
\scalebox{0.9}{ 
\begin{tabular}{|l|l|l|l|l|} 
\hline
Rank-List$_1$ & Rank-List$_2$ & $\rho-bbw$ & $\rho-ran$ & $\rho-full$ \\ \hline \hline
$UUR$         & Ground truth     & \textbf{0.67}    & \textbf{0.64}      & \textbf{0.62}             \\ \hline
$UTR$         & Ground truth	  &\textbf{0.66}      & \textbf{0.63}     & \textbf{0.63}             \\ \hline
$UPR$         & Ground truth      &\textbf{0.66}       &\textbf{0.64}    & \textbf{0.62}             \\ \hline
Baseline   & Ground truth      &0.49     &0.14        & 0.26                         \\ \hline
\end{tabular}
}
\end{table}

\subsection{Rank list alignment across rank ranges}
The number of \TG\textit{s} falling in each bucket across the three rank lists are same and are noted in table~\ref{tab:BucketWordDistribution}. Thus, the \textit{precision} and \textit{recall} as per the definition are also the same. The bucket-wise \textit{precision}/\textit{recall} for the baseline and $UUR$ with respect to the ground truth are noted in table~\ref{tab:BucketWisePR}. We observe that while in the SB bucket both the baseline and $UUR$ perform equally well, for all the other buckets $UUR$ massively outperforms the baseline. This implies that for the case where the likeliness of borrowing is the strongest, the baseline does as good as $UUR$. However, as one moves down the rank list, $UUR$ turns out to be a considerably better predictor than the baseline. The overall \textit{macro} and \textit{micro} \textit{precision}/\textit{recall} as shown in table~\ref{tab:OverallMiMaPrRc} further strengthens our observation that $UUR$ is a  better metric than the baseline.

\begin{table}
\centering
\caption{Number of words falling in each bucket of three bucket sets.}
\label{tab:BucketWordDistribution}
\scalebox{0.7}{ 
\begin{tabular}{|c|c|c|c|}
\hline
Bucket type & Ground truth bucket   & Baseline bucket  & $UUR$ bucket 		\\ \hline \hline
SB         & 11      	& 11 			& 11	\\ \hline
LB  	 	& 11      	& 11 			& 11    \\ \hline
BL         & 12      	& 12 			& 12    \\ \hline
LM        	& 11      	& 11 			& 11    \\ \hline
SM         	& 12      	& 12 			& 12    \\ \hline
\end{tabular} }
\end{table}

\begin{table}
\centering
\caption{Bucket-wise \textit{precision}/\textit{recall}. Best results are marked in bold.}
\label{tab:BucketWisePR}
\scalebox{0.8}{
\begin{tabular}{|c|c|c|}
\hline
Bucket type & \textit{prec.}/\textit{rec.} Baseline   & \textit{prec.}/\textit{rec.} $UUR$  		\\ \hline \hline
SB         & \textbf{0.27}     	& \textbf{0.27} 			                \\ \hline
LB  	 	& 0.09      	& \textbf{0.18}			                   \\ \hline
BL         & 0.08   		& \textbf{0.33} 			                   \\ \hline
LM        	& 0.18   		& \textbf{0.36} 			                   \\ \hline
SM         	& 0.33   		& \textbf{0.50} 			                   \\ \hline
\end{tabular}}
\end{table}
\begin{table}
\vspace*{-4mm}
\centering
\caption{  Overall \textit{macro} and \textit{micro} \textit{precision}/\textit{recall}. Best results are marked in bold.}
\label{tab:OverallMiMaPrRc}
\scalebox{0.8}{
\begin{tabular}{|c|c|c|}
\hline
Measure &  Baseline   &  $UUR$  		\\ \hline \hline
\textit{Macro prec.}/\textit{rec.}         & 0.19      	& \textbf{0.33} 			                \\ \hline
\textit{Micro prec.}/\textit{rec.}         & 0.19   			& \textbf{0.33}  			            \\ \hline
\end{tabular} }
\end{table}

\subsection{Age group based analysis}
\begin{figure}
\caption{Age distribution of the survey participants.}
\centering
\vspace{-3mm}
\includegraphics[width=0.5\textwidth]{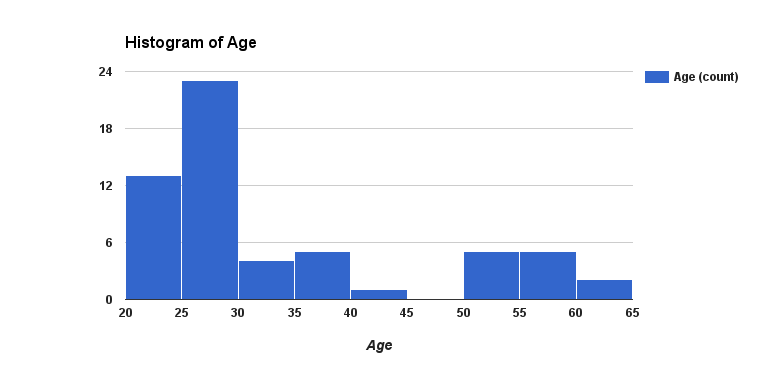}
\label{fig:AgeDistribution}
\end{figure}

As already discussed earlier, we split the ground truth responses based on the age group of the survey participants. In particular, we wish to observe the effect of this split on our ranking results. The distribution of participants of different age groups is shown in figure~\ref{fig:AgeDistribution}. Based on this distribution we split the responses into two groups as discussed earlier -- (i) young population (age $< 30$) and (ii) elderly population (age $\ge 30$).

\noindent{\bf Rank correlation}: The Spearman's rank correlation of $UUR$ and the baseline rank lists with these two ground truth rank lists are shown in table~\ref{tab:SpearManAcrossAge}. Interestingly, the correlation between $UUR$ rank list and the young population ground truth is better than the elderly population ground truth. This possibly indicates that $UUR$ is able to predict recent borrowings more accurately. However, note that the $UUR$ rank list has a much higher correlation with both the ground truth rank lists as compared to the baseline rank list.

\begin{table}
\centering
\caption{Spearman's rank correlation across the two age groups. Best results are marked in bold.}
\label{tab:SpearManAcrossAge}
\scalebox{0.8}{ 
\begin{tabular}{|l|l|l|}
\hline
Rank-List$_1$ & Rank-List$_2$ & $\rho$  \\ \hline \hline
Baseline   & Ground-truth-Young      & 0.26 \\ \hline
$UUR$         & Ground-truth-Young      & \textbf{0.62}  \\ \hline
Baseline   & Ground-truth-Elder      & 0.27 \\ \hline
$UUR$         & Ground-truth-Elder      & \textbf{0.53} \\ \hline
\end{tabular}
}
\end{table}

\noindent{\bf Rank ranges}: Table ~\ref{tab:NumWrdsYagOagGT} shows the bucket-wise \textit{precision} and \textit{recall} for $UUR$ and the baseline metrics with respect to two new ground truths. For the young population once again the number of words in each bucket for all the three sets is the same thus making the values of the \textit{precision} and the \textit{recall} same. In fact, the \textit{precision}/\textit{recall} for this ground truth is exactly same as in the case of the original ground truth. 

In contrast, when we consider the ground truth based on the responses of the elderly population, the number of words across the different buckets are different across the three sets. In this case, we observe that the \textit{precision}/\textit{recall} values are better for the $UUR$ metric in SB, LB and LM buckets. 

\begin{table}
\centering
\caption{Bucket-wise \textit{precision}/\textit{recall} for $UUR$ and the $baseline$ metrics for the two new ground truths. Best results are marked in bold.}
\label{tab:NumWrdsYagOagGT}
 \scalebox{0.8}{ 
\begin{tabularx}{0.58\textwidth}{| X | X | X | X | X | X | X |} 
\hline
Bucket type & Young-Baseline \textit{prec.}/\textit{Rec.} & Young-$UUR$ \textit{prec.}/\textit{rec.}  & Elder-Baseline \textit{prec.} & Elder-$UUR$ \textit{prec.} & Elder-baseline \textit{rec.} & Elder-$UUR$ \textit{rec.} 		\\ \hline \hline
SB         &\textbf{0.27}     &\textbf{0.27}  	&0.27     &\textbf{0.36}  	&0.25		&\textbf{0.33} \\ \hline
LB  	 	&0.09     &\textbf{0.18}  	&0.09     &\textbf{0.18}  	&0.08		&\textbf{0.17}	\\ \hline
BL         &0.08     &\textbf{0.33}		&\textbf{0.16}     &0.08  	&\textbf{0.28}		&0.14	\\ \hline
LM        	&0.18     &\textbf{0.36}		&0.18     &\textbf{0.45}  	&0.14		&\textbf{0.35}	\\ \hline
SM         	&0.33     &\textbf{0.5}   		&\textbf{0.41}     &0.25    	&\textbf{0.41}		&0.25  	\\ \hline
 
\end{tabularx} 
} 
\end{table} 

Finally, the overall  \textit{macro} and \textit{micro} \textit{precision} and \textit{recall} for both the age groups are noted in table~\ref{tab:OverallMiMaAge}. Once again, for both the young and the elderly population based ground truths, the \textit{macro} and \textit{micro precision} and \textit{recall} values for the $UUR$ metric are higher compared to that of the baseline. 

\begin{table}
\centering
\caption{Overall \textit{macro} and \textit{micro} \textit{precision} and \textit{recall} for the two new ground truths. Best results are marked in bold.}
\label{tab:OverallMiMaAge}
\scalebox{0.8}{ 
\begin{tabularx}{0.6\textwidth}{|X|X|X|X|X|}
\hline
Measure 					&  Young-Baseline    	&  Young-$UUR$ 		&  Elder-Baseline    	& Elder-$UUR$	 	\\ \hline \hline
\textit{Macro} \textit{prec.} 	&  0.19			    &  \textbf{0.33} 			&  0.22            	&  \textbf{0.27}			\\ \hline
\textit{Macro} \textit{rec.}  		&  0.19     			&  \textbf{0.33}			&  0.23				&  \textbf{0.25}			\\ \hline
\textit{Micro prec.}/\textit{rec.} 	&  0.19			&  \textbf{0.33} 			&  0.23	  	 	& \textbf{0.26}         	\\ \hline

\end{tabularx}
}
\end{table}

\subsection{Results based on the extent of language mixing}

As mentioned earlier, we divide the set of 3982 users into three categories. The Spearman's correlation between $UUR$ and the ground truth for each of these buckets are given in table~\ref{tab:UserSetBucket}. As we can see, for Low bucket the $\rho$ value is maximum. This points to the fact that the signals of borrowing is strongest from the users who rarely mix languages. 

\begin{table}
\centering
\caption{Spearman's correlation between $UUR$ and the ground truth in the different user buckets. Best results are marked in bold.}
\label{tab:UserSetBucket}
\scalebox{0.8}{ 
\begin{tabularx}{0.6\textwidth}{|X|X|X|}
\hline
Bucket &  Number of users &    $\rho$				 	\\ \hline \hline
High   &  	402          & 	 0.52		\\ \hline
Mid   &  	844          & 	 0.60		\\ \hline
Low   &  	2736         & 	\textbf{0.65}		\\ \hline 

\end{tabularx}
}
\end{table}

%% file: 60Conclusion.tex
\section{Conclusion}
In this paper we presented exhaustive experiments to quantify the likeliness of borrowing of a set of appropriately sampled \TG\textit{s}. Some of our key contributions are
\begin{compactitem}
\item We introduced a context based classification method to appropriately sample a list of 57 English \TG\textit{s}.  
\item We launched an extensive online survey among $58$ human judges to prepare a ground-truth data indicating the likeliness of borrowing of a \TG.
\item As a central contribution, we defined three closely related novel metrics based on social media signals indicating the likeliness that a \TG~shall be borrowed.  
\end{compactitem}

Some of our key results from this study are:
\begin{compactitem}
\item Remarkably, the rankings based on our metrics are more than two times more correlated to the ground-truth ranking compared to that of the baseline.
\item Further, our metrics predict the likeliness of borrowing better for the young population (age group below $30$) which possibly indicates that these are able to capture the early signals of borrowing.
\item Finally, if we compute the metrics from the tweets of those users who seldom code-mix, we obtain the best signals of borrowing.
\end{compactitem}

There are quite a few interesting directions of future research. First, we would like to obtain more theoretical insights into the better performance of these metrics by possibly having a dynamical model of word borrowing. Further, we would also like to incorporate our findings into standard tasks of multilingual IR and NLP and test if this leads to systematic performance enhancements. 